# Automated ICD Coding using Extreme Multi-label Long Text Transformer-based Models

**Leibo Liu[1*], Oscar Perez-Concha[1], Anthony Nguyen[2], Vicki Bennett[3], Louisa Jorm[1]**

[1]Centre for Big Data Research in Health, University of New South Wales, Sydney, Australia

[2]The Australian e-Health Research Centre, CSIRO, Brisbane, Queensland, Australia

[3]Metadata, Information Management and Classifications Unit (MIMCU), Australian Institute of Health and Welfare, Canberra, Australian Capital Territory, Australia

*Corresponding Author: Leibo Liu, Centre for Big Data Research in Health, Level 2, AGSM Building (G27) University of New South Wales Sydney, Kensington, New South Wales, 2052 Australia (z5250377@ad.unsw.edu.au, +61 2 9065 7847)

# ABSTRACT

**Background**: Encouraged by the success of pretrained Transformer models in many natural language processing tasks, their use for International Classification of Diseases (ICD) coding tasks is now actively being explored. In this study, we investigated three types of Transformer-based models, aiming to address the extreme label set and long text classification challenges that are posed by automated ICD coding tasks.

**Methods**: The Transformer-based model PLM-ICD achieved the current state-of-the-art (SOTA) performance on the ICD coding benchmark dataset MIMIC-III. It was chosen as our baseline model to be further optimised. XR-Transformer, the new SOTA model in the general extreme multi-label text classification domain, and XR-LAT, a novel adaptation of the XR-Transformer model, were also trained on the MIMIC-III dataset. XR-LAT is a recursively trained model chain on a predefined hierarchical code tree with label-wise attention, knowledge transferring and dynamic negative sampling mechanisms.

**Results**: Our optimised PLM-ICD model, which was trained with longer total and chunk sequence lengths, significantly outperformed the current SOTA PLM-ICD model, and achieved the highest micro-F1 score of 60.8%. The XR-Transformer model, although SOTA in the general domain, did not perform well across all metrics. The best XR-LAT based model obtained results that were competitive with the current SOTA PLM-ICD model, including improving the macro-AUC by 2.1%.

**Conclusion**: Our optimised PLM-ICD model is the new SOTA model for automated ICD coding on the MIMIC-III dataset, while our novel XR-LAT model performs competitively with the previous SOTA PLM-ICD model.

# INTRODUCTION

Clinical coding plays a crucial role in translating free text notes about diseases or procedures into a standardized code format.[1] The International Classification of Diseases (ICD) has been used to code and compare statistics on mortality and morbidity for more than a century.[2] Clinical modifications of ICD such as ICD-9-CM (USA), ICD-10-CM/PCS (USA), ICD-10-AM (Australia) and ICD-10-CA (Canada) have been developed by many countries for their national clinical coding.[3] Clinical coders conduct the clinical coding process and assign the relevant ICD codes to each patient episode of care according to the patient record and clinical notes.[4] Due to the expensive and labour-intensive manual coding workflow, there is a need to explore automated ICD coding to make the process more productive, efficient, accurate and consistent.[5] In addition, with the planned global implementation of ICD-11, which is a fully digitised tool, there will be even greater opportunities to automate the coding of clinical notes.[2]

In recent years, the rapid developments in the field of natural language processing (NLP) have led to a renewed interest in automated ICD coding.[6-12] From an NLP perspective, ICD coding is an extreme multi-label long text classification (XMLTC) task given the high number of labels to be classified (e.g., ICD-9-CM and ICD-10-CM/PCS have 17,849 and 141,747 codes respectively.[13]), and the length of clinical notes (e.g., the average length of discharge summaries of the well-known Medical Information Mart for Intensive Care (MIMIC-III)[14] dataset is ~1,855 words). Therefore, automated ICD coding requires the assignment of multiple codes from an extremely large code set to long sequences of text in clinical notes.

Many approaches have been proposed for automated ICD coding using different deep learning models such as Gated Recurrent Unit (GRU)[15 16], Convolutional Neural Network (CNN)[7 8 17-21] and Long-Short Term Memory (LSTM)[6 9]. With the emergence of Transformer models[22] and the superior performance of pretrained Transformer models in many NLP tasks[23-25], more and more researchers have begun to study Transformer-based models for ICD coding tasks.[10-12 26-30] However, two main challenges for fine-tuning pretrained Transformer models on the ICD coding tasks have not yet been fully solved:1) long text sequences and 2) extremely large code set.

The pretrained Transformer models BERT[23] and RoBERTa[24] restrict the maximum input sequence to 512 tokens due to the complexity increasing quadratically with the sequence length. Theoretically, XLNet[25] can handle any length of sequence. However, it needs vast computing resources for fine-tuning on long sequences because of the need to use all possible permutations of the factorization order. In order to solve the challenge of long text, our previous ICD coding study[11] proposed HiLAT, a hierarchical label-wise Transformer model that segmented the long text into multiple small chunks of up to 512 tokens each. The chunks were sequentially fed into the pretrained Transformer model for encoding the long text. Likewise, Huang et al. [12] used the same segmentation mechanism in their Transformer-based model PLM-ICD, which is the current state-of-the-art (SOTA) model for ICD coding on the MIMIC-III dataset. Another solution to the long text problem is to use Longformer[31] or BIGBIRD[32], which replace the standard full self-attention with a different sparse attention mechanism, so that both can handle up to 4,096 tokens, which is 8 times the maximum input sequence of BERT and RoBERTa. Therefore, there are two solutions to resolve the long text challenge for the ICD coding tasks: 1) Segment the long text into multiple small chunks. Each chunk can have up to 512 tokens per chunk. This solution can have any number of chunks and encode more than 4,096 tokens. 2) Truncate the long text to a fixed 4096 tokens and use Longformer or BIGBIRD to generate the token representations.

Some attempts have been made in other domains to address the challenge of an extremely large label space by fine-tuning pretrained Transformer models. Chang et al.[33] proposed X-Transformer, the first scalable method to fine-tune pretrained Transformer models on an extreme multi-label text classification (XMTC) problem. X-Transformer comprises three stages: cluster, matcher, and ranker. The cluster indexes the extreme label set semantically and groups $L$ labels into $K$ clusters. The matcher maps each text instance to the relevant clusters. The ranker assigns the subset of labels from the retrieved clusters to the text instances. Similarly, Jiang et al.[34] proposed LightXML which consists of label clustering, text representation, label recalling and label ranking. Unlike X-Transformer, which needs to train one model at each stage, LightXML provides an end-to-end training with a dynamic negative label sampling mechanism, which made the model focus on a small number of negative labels that were dynamically sampled for each training instance. Both X-Transformer and LightXML have only one level of clustering. This may lead to an extremely large number of clusters as they fix the size of the clusters to a small constant number of labels per cluster ($\leq 100$). To resolve this issue, Zhang et al.[35] proposed a novel recursive method, XR-Transformer, to reduce the computational complexity through recursively fine-tuning the pretrained Transformer models on a series of easy-to-hard sub-tasks which are derived from the original XMTC task. XR-Transformer first builds a hierarchical label tree (HLT)[36] from the original label space and then recursively trains the models on each level of HLT using the dynamic negative label sampling mechanism, similar to LightXML. XR-Transformer achieved better or similar SOTA performances on six public XMTC datasets with significantly less training time. However, these methods were evaluated on short text with 128 or 512 tokens to reduce the usage of the GPU memory. To the best of our knowledge, none of the previous research has applied these methods on XMLTC problems such as the ICD coding tasks.

In this study, we explored and evaluated three different Transformer-based model architectures for ICD coding on the MIMIC-III dataset, with the aim of addressing the challenges of long text sequences and extremely large code set. Our contributions are as follows:

- We chose the current SOTA model, PLM-ICD, as our baseline model and further optimised it by training it with longer sequence lengths and larger chunk numbers. It was hypothesised that using longer sequence lengths and/or a larger number of chunks would better support the long text data characteristic of the ICD coding tasks. We also investigated different pretrained Transformer models in PLM-ICD.
- We adapted the original XR-Transformer (extreme multi-label) model to support long sequences as input, using a BIGBIRD-based encoder.
- We proposed XR-LAT, a novel recursive label-wise attention Transformer-based model which was motivated by the design of XR-Transformer to support extreme multi-labels and the robustness of label-wise attention on ICD coding tasks[7 9 11 12]. XR-LAT was recursively trained on the natural ICD code structure with a knowledge transferring and dynamic negative sampling mechanism. XR-LAT employed a segmentation mechanism to handle long text.
- Similar to Li et al.'s work[37], we further pretrained BIGBIRD on all the clinical notes of MIMIC-III using different settings, named this model ClinicalBIGBIRD, and evaluated its impact on the ICD coding tasks.

## MATERIALS AND METHODS

**Datasets**

MIMIC-III, a publicly accessible benchmark database, is widely used by ICD coding researchers. It contains data for 46,520 patients and 15 types of clinical notes.[14] Similar to previous studies,[7 9 11 12] we used the hospital discharge summaries of MIMIC-III for ICD coding. As we focused on the extreme large code set, we chose to maximise the number of ICD-9-CM codes by using the MIMIC-III-full dataset. Following previous studies[7 9], the same pre-processing was applied to the discharge summaries. We removed some special characters (e.g., ==, --, __) and the de-identification surrogates (e.g., [**First Name8 (NamePattern2) **], [**2151-7-16**], [**Hospital 1807**]) from all the discharge summaries. Table 1 shows the descriptive statistics of MIMIC-III-full dataset. The counts of training, validation, and test sets were based on the patient hospital admissions, of which one patient could have one or more. In addition, the MIMIC-III-CN[11] dataset, which consists of all the clinical notes such as nursing notes, discharge summaries, radiology, and ECG reports, was used for continually pretraining BIGBIRD.

*Table 1. Descriptive statistics of MIMIC-III-full dataset.*

| Item | Count |
| --- | --- |
| Training set | 47,719 |
| Validation set | 1,631 |
| Testing set | 3,372 |
| Unique ICD-9-CM codes | 8,929 |
| Min. / Median / 75th percentile / 95th percentile / Max. words per document | 52 / 1,701 / 2,345 / 3,560 / 10,497 |
| Min. / Median / 75th percentile / 95th percentile / Max. tokens per document* | 61 / 2,405 / 3,353 / 5,110 / 17,662 |
| Min./ Avg / Max. codes per document | 1 / 16 / 71 |

*\* The counts of tokens were calculated using BERT tokenizer. The counts can vary depending on different pretrained Transformer models.*

**Pretrained Transformer models**

In the biomedical domain, many domain-specific Transformer models have been continually pretrained or pretrained from scratch using the collection of PubMed[1] biomedical abstracts and/or clinical notes from MIMIC-III, including ClinicalBERT[38], PubMedBERT[39], ClinicalplusXLNet[11] and RoBERTa-PM-M3[40]. The domain-specific pretrained Transformer models have been proven to perform better than those pretrained on general-domain data in many biomedical NLP tasks.[11 39-41] RoBERTa-PM-M3 achieved the best performance on the multi-label document classification task compared to other domain-specific Transformer models.[40] ClinicalplusXLNet performed better than ClinicalBERT and PubMedBERT on the MIMIC-III top 50 ICD coding task.[11] Therefore, we chose both RoBERTa-PM-M3 and ClinicalplusXLNet to be fine-tuned in this study for ICD coding.

Motivated by the success of domain-specific pretraining, we further pretrained BIGBIRD on the MIMIC-III-CN dataset and named this model ClinicalBIGBIRD. We first pre-processed the MIMIC-III-CN dataset to split the clinical notes with more than 4,094 tokens into multiple chunks of up to 4,094 tokens. The HuggingFace Transformer library[2] was used to pretrain BIGBIRD from the checkpoint of google/bigbird-roberta-base[3] using the hyperparameters (maximum sequence length 4,096, batch size 64, warmup steps 10,000, single precision floating point (32 bits – FP32) for 500,000 steps (details in the Section 1 of Supplementary Material). Li et al. also pretrained Clinical-BigBird on the same corpus.[37] However, they used half-precision floating point format (16 bits - FP16) to accelerate training during the pretraining process and pretrained the model for 300,000 steps with a batch size of 12 and a different BIGBIRD checkpoint. Both ClinicalBIGBIRD and Clinical-BigBird were fine-tuned during ICD coding training to take long sequences as input.

**Baseline models**

The PLM-ICD model achieved the latest SOTA performance on the MIMIC-III-full dataset.[12] It contains a Transformer layer with a segment pooling mechanism, a label-wise attention layer and a classifier layer. The segment pooling mechanism resolves the long text challenge of fine-tuning pretrained Transformer models. The whole document $D$, which has $t$ tokens, is split into $s$ consecutive chunks. Each chunk has $c$ tokens where $c \leq 512$ and the total sequence length is $z = c \times s$, if $t < z$, the padding tokens are inserted, if $t > z$, the document is truncated. Each chunk is encoded by the Transformer layer separately. The hidden representations of these chunks are concatenated to obtain document representation $H = [h_1, h_2, ..., h_z]$, where $h_i$ is the hidden representation for token $i$. The goal of the label-wise attention layer is to transform document representation $H$ to label-specific document representation $d_j$, where $j$ is the $j^{th}$ code of the ICD code set. $d_j$ is input into the classifier layer to compute the probability of the $j^{th}$ code. The binary cross-entropy (BCE) loss function is used in the training process.

---

[1] https://pubmed.ncbi.nlm.nih.gov/
[2] https://github.com/microsoft/huggingface-transformers
[3] https://huggingface.co/google/bigbird-roberta-base

In this study, we used PLM-ICD as our baseline model and conducted the experiments using different total sequence length $z$, chunk number $s$ and chunk size $c$. Four pretrained Transformer models (the base version of RoBERTa-PM-M3, ClinicalplusXLNet, ClinicalBIGBIRD, and Clinical-BigBird) were used in the Transformer layer respectively. Finally, we also tried another loss function, Asymmetric Loss (ASL)[42 43] that has been proven to outperform binary cross-entropy loss for multi-label classification tasks.

**XR-Transformer with BIGBIRD-based encoder**

The XR-Transformer model[4] implementation only supports BERT, RoBERTa and XLNet models as the encoder. The long text must be truncated up to 512 tokens before feeding into the model. Therefore, we implemented a BIGBIRD-based encoder which can encode up to 4,096 tokens at a time. We trained XR-Transformer with ClinicalBIGBIRD as the encoder using the hyperparameters: truncate length 4,096, batch size 32, max leaf size 16, min codes 16[5], max steps 10,000, warmup steps 1,000 and save steps 500.

**XR-LAT model**

Inspired by the success of XR-Transformer on the XMTC problem [35] and the robustness of label-wise attention mechanism on ICD coding tasks [7 9 11 12], we proposed XR-LAT, a recursively trained model chain on predefined hierarchical code tree. Figure 1 illustrates the architecture of XR-LAT model which sequentially trained four sub-models in a waterfall manner. Each sub-model used the same architecture as the baseline model PLM-ICD. We used bootstrapping to transfer the knowledge learnt from parent level to children level and dynamic negative sampling to make the training from 'easy' top level with few codes to 'hard' bottom level with an extreme large code set. The subsequent sections describe XR-LAT in more detail.

---

[4] https://github.com/amzn/pecos/tree/mainline/pecos/xmc/xtransformer
[5] https://github.com/amzn/pecos/tree/mainline/examples/xr-transformer-neurips21/params/eurlex-4k/roberta

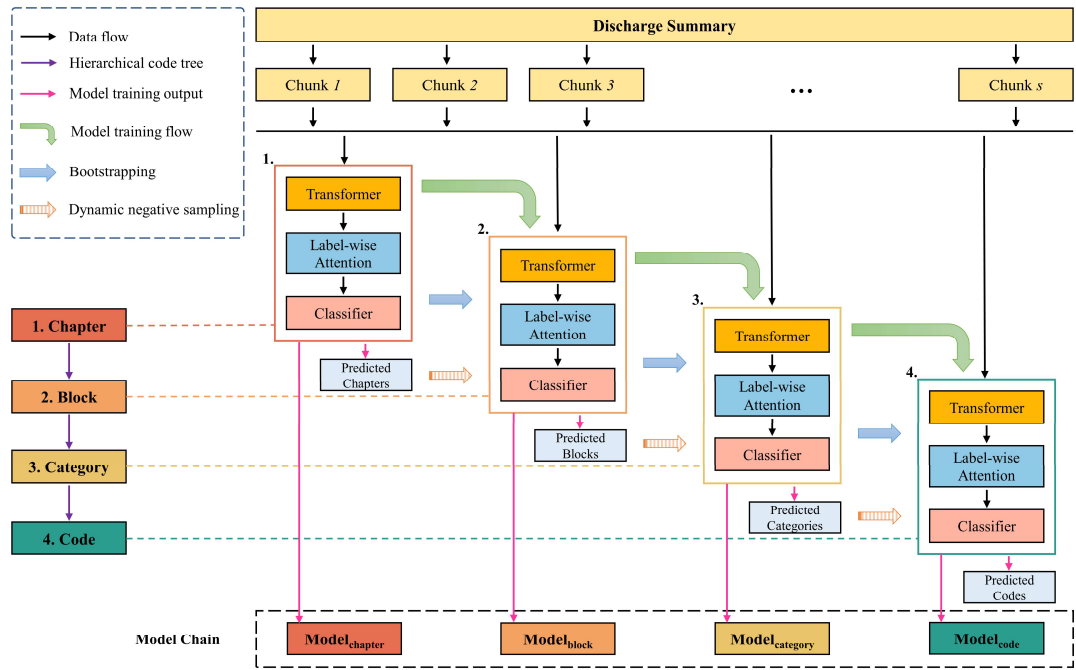

*Figure 1. Architecture and training process of XR-LAT model.*

**Hierarchical code tree (HCT)**

According to the tabular structure[6] of ICD-9-CM, we designed a four-level HCT including Chapter, Block, Category and Code levels. We built HCT of depth four on the MIMIC-III-full dataset into a series of indexing matrices $\{T^{(k)}\}_{k=1}^{4}$, where $T^{(k)} \in \{0, 1\}^{V_k \times V_{k-1}}$. There were 36 chapters, 279 blocks, 1,167 categories and 8,929 codes in the HCT. Therefore, $V_0 = 1, V_1 = 36, V_2 = 279, V_3 = 1,167 \ and \ V_4 = 8,929$, where $V_i$ was the code number of *i* level and $V_0$ always equalled to one. During the training process, the instance label set of the current level was calculated based on the label set of its children level. The instance label set of Code level was the label set of the training data. For example, $L^4 \in \{0, 1\}^{N \times 8,929}$ denoted the label set of Code level, where N was the instance number of the training data. The label set of Category level was then calculated using $L^3 = L^4 T^4$. Figure 2 shows a part of the built HCT of the MIMIC-III-full dataset.

---

[6] http://icd9.chrisendres.com/index.php?action=contents

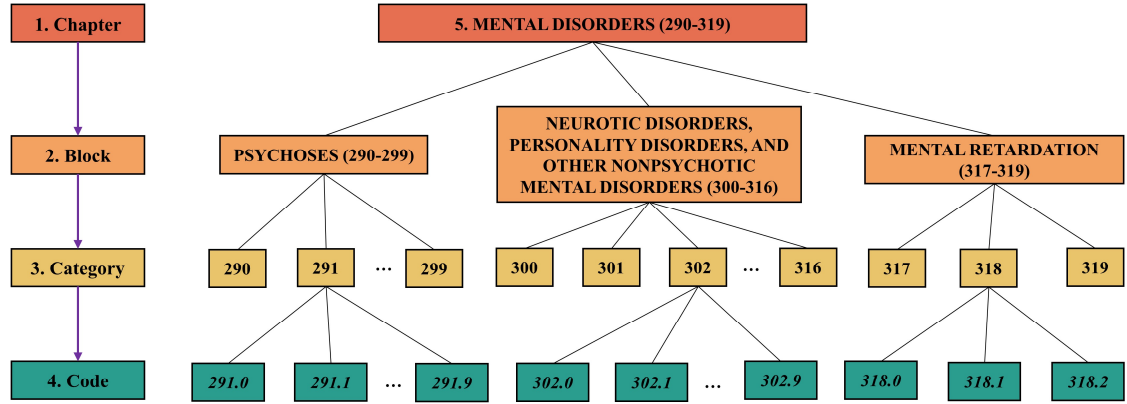

*Figure 2. A part of hierarchical code tree of MIMIC-III-full dataset.*

**Bootstrapping**

Hierarchical knowledge transferring has been demonstrated to benefit the training of machine learning models.[35 43] In XR-LAT, we employed this mechanism to bootstrap the model weight initialization of the current level from the trained model weights of its parent level. Given that, $W_t^1, W_{la}^1$, and $W_{cl}^1$ were the final weights of Transformer layer (*t*), label-wise attention layer (*la*) and classifier layer (*cl*) of the Chapter level, respectively. Instead of randomly initializing the weights of the Block level, we used $W_t^1$ as the initial weights $W_t^2$ of the Transformer layer and calculated $W_{la}^2 = T^2 W_{la}^1$ and $W_{cl}^2 = T^2 W_{cl}^1$. This method initialized the weights of all the sibling codes to be the same in both the label-wise attention and classifier layer. We named this method as BootstrapEqual. However, it may cause misleading information because the sibling codes were not identical. Following the work of Ren et al.[43], we investigated the hyperbolic correction method for bootstrapping process. Hyperbolic correction was to use hyperbolic embeddings[44] of HCT to dynamically calculate the weights for the sibling codes. Hyperbolic embeddings can be better at capturing the hierarchical representations on hyperbolic space which outperforms the embeddings on the Euclidean space.[44] Firstly, we trained a model based on the HCT $\{T^{(k)}\}_{k=1}^4$ using gensim library[7]. Secondly, the hyperbolic embeddings of all the tree nodes were extracted from the model as $\{E^{(k)}\}_{k=1}^4$, where $E^{(k)} \in \{0,1\}^{V_k \times 50}$ (50 was the predefined vector size). Finally, we calculated the weights of the label-wise attention layer using $W_{la}^2 = T^2 W_{la}^1 + f(E^2)$, where $f(E^2)$ was a fully connected layer. This method was named as BootstrapHyperC. We studied these two different bootstrapping methods in our experiments.

**Dynamic negative sampling**

Dynamic negative label sampling has already been used in XMTC problems.[34 35] It makes the model focus on a subset of all the labels for each instance during the training process to reduce the computational complexity. In XR-LAT, the negative labels were sampled dynamically according to the predicted results and the true labels of the parent level for each instance in the training set. For example, $p_i^1, y_i^1 \in \mathbb{R}^{36}$ were the predicted labels and the true labels for the instance $i$ in the Chapter level, respectively. The model on the Block level was trained on the label subset $m_i^2 = T^2 binary(p_i^1 + y_i^1)$ for the instance $i$. As dynamic negative label sampling could only be used during training, at inference time, only the codes whose parent codes were predicted as positive are sampled, where $m_i^2 = T^2 binary(p_i^1)$.

---

[7] https://radimrehurek.com/gensim/models/poincare.html

**Model evaluation**

Following previous works[7 9 12], we evaluated the models using the following metrics: macro-/micro- AUC (Area Under the ROC curve), macro-/micro- F1 scores, and precision at K (P@5, P@8, P@15). Macro-AUC or macro-F1 is the mean of the AUC or F1 of each code and treats all codes (common and rare codes) as equally important. Micro-AUC or micro-F1 is based on the sum of all true positives (TP), false positives (FP), and false negatives (FN) over all the codes and is potentially biased towards the common codes in imbalanced datasets. P@k is the average of the precision of the top-k predicted codes.

All the models were trained on a NVIDIA A100 GPU (80GB). We describe more details about model training in Section 2 of the Supplementary Material.

# RESULTS

**Hyperparameters**

We conducted a manual search of hyperparameters by training six baseline models using different combinations, as shown in Supplementary Table S1. The hyperparameters for this study were batch size at 8, learning rate at 5e-5, weight decay at 0.1, dropout at 0.1, warmup steps at 5,000, and training steps at 60,000. The random seed was set to 2022. The logging and save steps were both 5,000. The mean and standard deviation of the weight normalization were fixed to 0.0 and 0.03 respectively when initializing the model linear layers.

**Baseline model results**

We performed ten experiments of the baseline model PLM-ICD using different chunk numbers, chunk sequence lengths, and pretrained Transformer models. The total sequence lengths of the experiments were 3,072, 4,096, 5,120 and 6,144. The main pretrained Transformer model we used in the experiments was RoBERTa-PM-M3 as it has been proven to outperform other domain-specific BERT variants[12]. To compare with other types of pretrained Transformer models, we also used

ClinicalplusXLNet, ClinicalBIGBIRD, and Clinical-BigBird in the experiments. Table 2 shows the experimental results on the MIMIC-III-full dataset. First, we reproduced the training of PLM-ICD model using the same settings as the original PLM-ICD study in the experiment BL-1. The scores were similar to the ones of the original study except the micro-F1 score which decreased by 1%. This may be caused by the random seed and different hyperparameters such as weight decay, dropout rates, and the weight normalization parameters of the linear layers. Next, we compared the performances of models using different total sequence lengths by varying the number of chunks while constraining the chunk sequence length to 128 (BL-2, BL-3, and BL-4). The model BL-3 with 40 chunks and a total sequence length of 5,120 achieved the best micro-F1 of 60.0%, followed by the model BL-2 with 32 chunks. After that, the model BL-5, which used the same settings as BL-3 except a longer chunk sequence length of 512, further improved the performance of BL-3 on each metric. Derived from BL-5, the model BL-6 with ASL loss function increased BL-5 macro-F1 by 1.1% but decreased on the other metrics, and the model BL-7 using ClinicalplusXLNet improved BL-5 macro-F1 by 0.5% and macro-AUC by 0.2%. Until now, all the previous experiments used the segmentation mechanism to handle long text. ClinicalBIGBIRD and Clinical-BigBird which support up to 4,096 tokens within a single chunk to tackle the long text problem was also investigated in the model BL-8 and BL-9. BL-8 performed better than BL-9 on F1 scores. Compared to the current best baseline model BL-5 which had a longer total sequence length of 5,120, the model BL-8 was suboptimal across all the metrics, especially macro-F1 by 2%. To verify this effect, we performed an additional two experiments BL-10 and BL-11 which used the segmentation mechanism but with the total sequence length of 4,096 (i.e., the same as ClinicalBIGBIRD in BL-8) and chunk sequence length of 512. The model BL-10 using RoBERTa-PM-M3 improved macro-F1 by 1.8% and micro-F1 by 1.2%, compared to BL-8. Similarly, the model BL-11 using ClinicalplusXLNet performed better than BL-8 on F1 scores. Overall, the model BL-5 performed the best in all the baseline models on micro-AUC, micro-F1 and P@k scores and significantly outperformed the current SOTA model BL-1.

*Table 2. Results of baseline model PLM-ICD on MIMIC-III-full dataset. The **bold** indicates the best result for each metric. BL-5 results were the average of the results performed using five different random seeds (see Supplementary Table S2).*

| Experiment No. | Total Sequence Length | Chunks | Chunk Sequence Length | Loss Function | Pretrained Transformer Model | AUC | | F1 | | P@k | | |
|---|---|---|---|---|---|---|---|---|---|---|---|---|
| | | | | | | Macro | Micro | Macro | Micro | P@5 | P@8 | P@15 |
| PLM-ICD[†] | 3,072 | 24 | 128 | BCE | RoBERTa-PM-M3 | 92.6 | 98.9 | 10.4 | 59.8 | 84.4 | 77.1 | 61.3 |
| BL-1[‡] | 3,072 | 24 | 128 | BCE | RoBERTa-PM-M3 | 92.5 | 98.9 | 10.4 | 58.8 | 84.1 | 76.4 | 61.0 |
| BL-2 | 4,096 | 32 | 128 | BCE | RoBERTa-PM-M3 | 92.9 | 98.9 | 10.8 | 59.8 | 84.7 | 77.4 | 61.8 |
| BL-3 | 5,120 | 40 | 128 | BCE | RoBERTa-PM-M3 | 92.9 | 98.9 | 10.3 | 60.0 | 84.7 | 77.4 | 61.9 |
| BL-4 | 6,144 | 48 | 128 | BCE | RoBERTa-PM-M3 | **93.3** | **99.0** | 11.4 | 59.5 | 84.2 | 77.0 | 61.7 |
| BL-5 | 5,120 | 10 | 512 | BCE | RoBERTa-PM-M3 | 93.1* ± 0.2 | **99.0*** ± 0.0 | 11.1* ± 0.1 | **60.7*** ± 0.1 | **85.2*** ± 0.1 | **78.0*** ± 0.1 | **62.4*** ± 0.1 |

| BL-6 | 5,120 | 10 | 512 | ASL | RoBERTa-PM-M3 | 92.6 | 98.9 | **12.1** | 59.5 | 84.4 | 77.0 | 61.8 |
| BL-7 | 5,120 | 10 | 512 | BCE | ClinicalplusXLNet | **93.3** | **99.0** | 11.5 | 60.4 | 84.0 | 77.0 | 61.9 |
| BL-8 | 4,096 | 1 | 4,096 | BCE | ClinicalBIGBIRD | 92.3 | 98.9 | 9.0 | 59.2 | 84.4 | 76.9 | 61.2 |
| BL-9 | 4,096 | 1 | 4,096 | BCE | Clinical-BigBird | 92.6 | 98.9 | 8.9 | 59.0 | 84.4 | 77.3 | 61.3 |
| BL-10 | 4,096 | 8 | 512 | BCE | RoBERTa-PM-M3 | 92.8 | 98.9 | 10.8 | 60.4 | 84.9 | 77.7 | 62.1 |
| BL-11 | 4,096 | 8 | 512 | BCE | ClinicalplusXLNet | 93.0 | 98.9 | 11.6 | 59.9 | 84.0 | 76.6 | 61.4 |

† indicates the results from the original study of PLM-ICD. ‡ Indicates the experimental results we reproduced using the same chunk number and sequence length as PLM-ICD study. * indicates that the improvement was significant with p < .05, compared to BL-1 (see Supplementary Table S3).

### XR-Transformer and XR-LAT results

To investigate extreme multi-label abilities of XR-Transformer on the ICD coding tasks, we trained XR-Transformer with BIGBIRD-based encoder using a long input sequence of 4,096. We also explored XR-LAT and two different variants of XR-LAT: XR-LAT-BootstrapEqual and XR-LAT-BootstrapHyperC. The two variants were ablation studies of XR-LAT, which did not use the dynamic negative sampling mechanism and only transferred the learnt knowledge in the training process. We compared XR-LAT and its variants to the baseline PLM-ICD models and XR-Transformer. We used ClinicalBIGBIRD to encode a long sequence of 4,096 in order to compare XR-LAT with XR-Transformer. Furthermore, we performed the experiments of XR-LAT and its two variants using the same settings as the current SOTA model BL-1 and the best baseline model BL-5 to investigate the chunk segmentation approach to handling long text. Table 3 reports the evaluation results of XR-Transformer and XR-LAT-based models. All the scores of XR-Transformer (XR-1) and XR-LAT (XR-2, XR-3, and XR-4) were suboptimal compared to the two baseline models (BL-1 and BL-5). The XR-LAT model XR-2 improved each metric at a large margin, compared to the XR-Transformer model XR-1. XR-LAT-based models (including XR-LAT-BootstrapEqual and XR-LAT-BootstrapHyperC) all performed better with longer sequence lengths. Furthermore, all the XR-LAT variant models (from XR-5 to XR-8) achieved higher macro-AUC and nearly the same micro-AUC, compared to the two baseline models (BL-1 and BL-5). The model XR-6 was the best model in all the XR type models. It did not perform better than the best baseline model BL-5 on F1 and P@k scores. However, it significantly improved macro-AUC, micro-AUC, and macro-F1, compared to the current SOTA model BL-1 (see Supplementary Table S4).

Table 3. Results of XR-Transformer and XR-LAT on the MIMIC-III-full dataset. The **bold** indicates the best result for each metric among the XR experiments. BL-1 is the current SOTA model and BL-5 is the best baseline model in this study.

| Experiment No. | Model Architecture | Total Sequence Length | Chunks | Chunk Sequence Length | Pretrained Transformer Model | AUC | | F1 | | P@k | | |
|---|---|---|---|---|---|---|---|---|---|---|---|---|
| | | | | | | Macro | Micro | Macro | Micro | P@5 | P@8 | P@15 |
| BL-1 | PLM-ICD | 3,072 | 24 | 128 | RoBERTa-PM-M3 | 92.5 | 98.9 | 10.4 | 58.8 | 84.1 | 76.4 | 61.0 |

| | | | | | | | | | | | |
|---|---|---|---|---|---|---|---|---|---|---|---|
| BL-5 | PLM-ICD | 5,120 | 10 | 512 | RoBERTa-PM-M3 | 93.1 | 99.0 | 11.0 | 60.8 | 85.3 | 78.0 | 62.5 |
| XR-1 | XR-Transformer | 4,096 | 1 | 4,096 | ClinicalBIGBIRD | 56.7 | 71.6 | 6.2 | 40.9 | 69.7 | 61.1 | 46.0 |
| XR-2 | XR-LAT | 4,096 | 1 | 4,096 | ClinicalBIGBIRD | 71.1 | 82.7 | 7.3 | 52.9 | 77.1 | 70.1 | 54.7 |
| XR-3 | | 3,072 | 24 | 128 | RoBERTa-PM-M3 | 72.1 | 82.3 | 8.3 | 53.6 | 76.7 | 69.8 | 54.9 |
| XR-4 | | 5,120 | 10 | 512 | RoBERTa-PM-M3 | 74.0 | 84.1 | 8.8 | 56.1 | 78.7 | 71.7 | 56.7 |
| XR-5 | XR-LAT-BootstrapEqual | 3,072 | 24 | 128 | RoBERTa-PM-M3 | 94.0 | 98.9 | 10.0 | 56.1 | 81.0 | 73.7 | 58.6 |
| XR-6 | | 5,120 | 10 | 512 | RoBERTa-PM-M3 | **94.6** | **99.0** | 10.8 | **58.3** | **82.1** | **74.9** | **59.9** |
| XR-7 | XR-LAT-BootstrapHyperC | 3,072 | 24 | 128 | RoBERTa-PM-M3 | 93.7 | 98.9 | 10.9 | 56.4 | 80.4 | 73.6 | 58.5 |
| XR-8 | | 5,120 | 10 | 512 | RoBERTa-PM-M3 | 94.3 | 98.9 | **11.3** | 58.1 | 81.3 | 74.4 | 59.6 |

**Training time**s

Figure 3 compares the training times for the different experiments that we carried out in this study. We excluded the model BL-9 from the comparison as its training time was nearly the same as the one of BL-8. In general, the training time depended on three factors: 1) Total sequence length: The training time increased along with increasing total sequence length given a fixed chunk sequence length and the same pretrained Transformer model (Figure 3a). 2) Chunk sequence length: The training time also increased with the chunk sequence length given a fixed sequence length and pretrained Transformer model (Figure 3b and Figure 3c). 3) Pretrained Transformer model type: The experiments (BL-7 and BL-11) with ClinicalplusXLNet needed three times longer than the experiments (BL-5 and BL-10) with RoBERTa-PM-M3 given the same sequence settings (Figure 3b and Figure 3c). Furthermore, the models that fine-tuned ClinicalBIGBIRD with a single long sequence of 4,096 had the longest training time, with nearly 105 hours for XR-LAT and 55 hours for XR-Transformer (Figure 3d). As shown in Figure 3e and Figure 3f, the XR-LAT-based models needed much longer training time than the baseline PLM-ICD models as they comprised of four training processes.

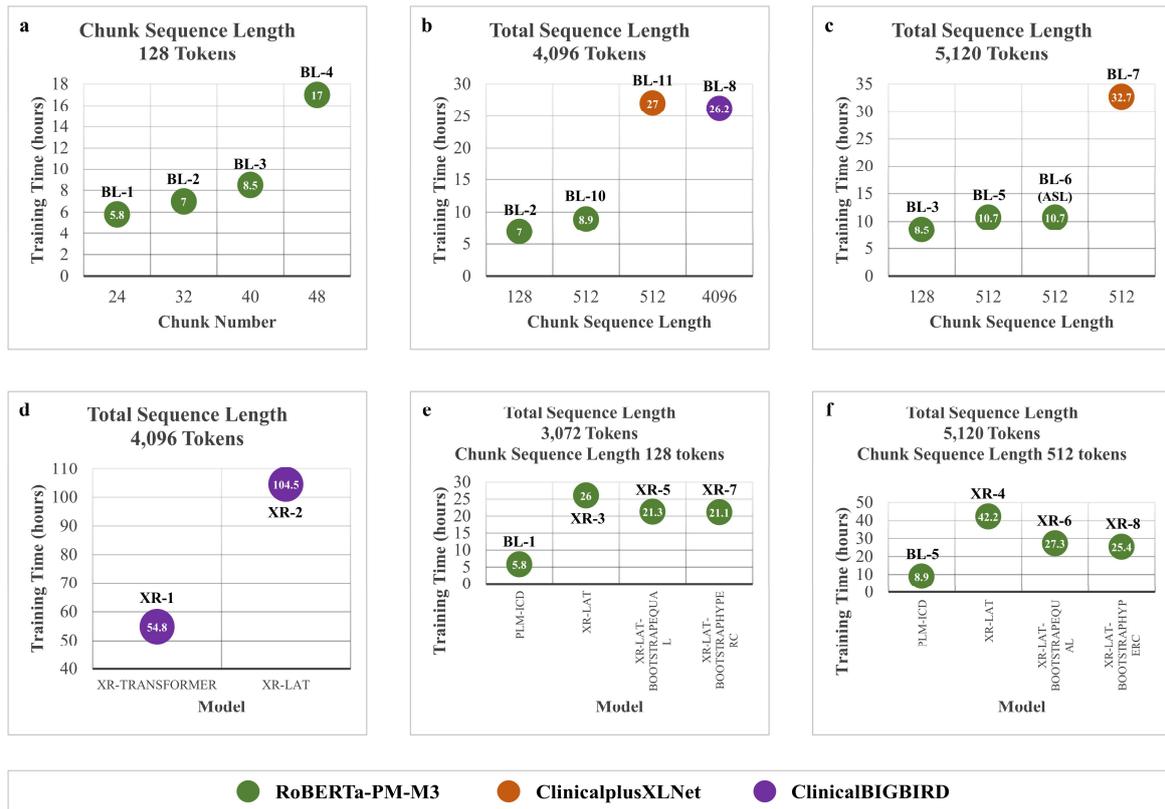

*Figure 3. Training time comparison. Different solid colour circles stand for different types of Transformer models. Green, orange, and purple are for RoBERTa-PM-M3, ClinicalplusXLNet, and ClinicalBIGBIRD, respectively. a) The chunk sequence length was fixed at 128. The chunk number increased from 24 to 48. b) The total sequence length was fixed at 4,096. The chunk sequence lengths were selected from 128, 512, and 4,096. BL-8 did not use the segmentation mechanism. c) The total sequence length was fixed at 5,120. The chunk sequence lengths were selected from 128 and 512. BL-6 used ASL loss function. d) The total sequence length was 4,096. No segmentation mechanism was used in the two experiments. e) The settings were total sequence length at 3,072 and chunk sequence length at 128. f) The settings were total sequence length at 5,120 and chunk sequence length at 512.*

## DISCUSSION

This study investigated three different Transformer-based model architectures for automated ICD coding, including one novel proposed approach, XR-LAT, that recursively fine-tuned the pretrained Transformer model on a hierarchical code tree with a label-wise attention layer and a label-wise feed-forward neural network (FFNN) layer using bootstrapping and dynamic negative label sampling mechanisms. The models were trained using MIMIC-III-full dataset to predict a subset of ICD-9-CM codes from a large set of 8,929 codes on discharge summaries with a 95[th] percentile of 3,560 words.

The selected baseline model PLM-ICD outperformed other ICD coding approaches and is the latest SOTA approach for the ICD coding tasks on the MIMIC-III-full dataset.[12] However, our work shows that it was not fully optimised. Our best baseline model BL-5, which used longer total sequence length of 5,120 and chunk sequence length of 512, significantly outperformed the current SOTA PLM-ICD model (BL-1). Our proposed model XR-LAT (XR-2) performed better than the XR-

Transformer model (XR-1) by a large margin using the same pretrained ClinicalBIGBIRD with a long sequence of 4,096 tokens. However, its performance was worse on all metrics, especially for macro-/micro- AUC, compared to the current SOTA PLM-ICD model BL-1 and our best baseline model BL-5. This may be caused by inference errors from the parent levels that propagate to the children levels in the hierarchical code tree. However, the XR-LAT variant models using bootstrapping produced better performance on macro-AUC (XR-6) and macro-F1 (XR-8), compared to the current SOTA model BL-1 and our best baseline model BL-5. This indicated that the bootstrapping paid more attention to infrequent codes.

To tackle the long text challenge of fine-tuning Transformer models, we investigated two different approaches in this study: 1) Segment long text to multiple short chunks; 2) Employ the Transformer models that are specifically designed for long sequences such as BIGBIRD. Our experiments showed that the former performed better than the latter if using the same total sequence length. In addition, the training time of the first approach was much faster than the other. Although BIGBIRD can take a long sequence as the input, it still has a maximum sequence length of 4,096. This will restrict the model performance on the datasets where there are many long texts with the most important information located at the end of the document. For example, the discharge diagnosis, which is the most important information for ICD coding, is at the end of the discharge summaries of the MIMIC-III-full dataset.[11] The segmentation approach can deal with sequences of any length to address this limitation. According to our experiments, the best total sequence length was 5,120 which was longer than the maximum sequence of BIGBIRD, and the chunk sequence length of 512 always performed better than 128 for RoBERTa-PM-M3 based models. However, the total sequence length could be a hyperparameter which varies between different datasets. In the MIMIC-III-full dataset, the 95$^{th}$ percentile for discharge summary length is 5,110 tokens, and less than 5% of discharge summaries needed to be truncated to 5,120 tokens during training.

In the general domain, the XR-Transformer approach has established the new SOTA results for extreme multi-label tasks on public XMTC datasets. However, it performed worse than the other two Transformer-based approaches investigated on the MIMIC-III-full dataset. Our proposed XR-LAT model XR-2 improved the macro-AUC by 14.4%, micro-AUC by 11.1%, macro-F1 by 1.1%, micro-F1 by 12.0%, P@5 by 7.5%, P@8 by 9.0%, and p@15 by 8.6%, compared to the XR-Transformer model XR-1. The label-wise attention mechanism, which represented the document differently for each code, may be positively contributing towards this improvement.[11]

In our experiments, we also explored the ASL loss function in the baseline model and hyperbolic correction in the XR-LAT variant. Both improved the macro-F1 by a large margin. This demonstrated that ASL and hyperbolic correction emphasized rare codes during the training process. They could potentially achieve better performance on datasets with an extreme long tail distribution.

**Limitations**

The present study has two limitations. All our experiments were conducted on a single dataset (MIMIC-III-full). The dataset includes 8,929 ICD-9-CM codes, which is much fewer than the full code set of ICD-10-CM/PCS (141,747 codes). There is a need to evaluate Transformer-based approaches on a comprehensively coded health dataset and/or publicly available XMTC datasets. The segmentation method in this study split the long text consecutively into small chunks. This may potentially harm model performance because the context was broken up at the edge of each chunk. One possible solution is to apply overlapping on adjacent chunks. The number of overlapping tokens is another hyperparameter that needs to be studied in the future.

# CONCLUSION

In conclusion, we compared three Transformer-based model architectures for automated ICD coding on the MIMIC-III-full dataset, with the aim of addressing two main challenges of fine-tuning pretrained Transformer models for automated ICD coding tasks: 1) long text and 2) extreme large code set. The optimised PLM-ICD baseline model BL-5 achieved the best performance with the total sequence length of 5,120. The model XR-Transformer, which is the new SOTA in the general XMTC domain, did not perform better than the baseline on the MIMIC-III-full dataset. Our novel proposed XR-LAT obtained much better performance than XR-Transformer, and the variant of XR-LAT which used bootstrapping produced competitive results to the current SOTA PLM-ICD model. Overall, Transformer-based models achieved the best performance on the automated ICD coding task. The segmentation method gave Transformer-based models the ability to tackle the long text challenge of ICD coding. However, further research should be undertaken to investigate the ability of Transformer-based to deal with the extreme large label set on datasets which include a greater number of codes than the MIMIC-III-full dataset.

# DATA AVAILABILITY

The source code of this study is available at https://github.com/leiboliu/xr-lat. And the MIMIC-III v1.4 can be downloaded via https://physionet.org/content/mimiciii/1.4/.

# AUTHOR CONTRIBUTIONS

LL designed and conducted all the experiments and wrote the manuscript. OP-C, AN, VB and LJ provided the supervision and suggestions and reviewed the manuscript.

# FUNDING

This study was supported by the Australian government and the Commonwealth Industrial and Scientific Research Organisation (CSIRO) through Australian Government Research Training Program scholarship and CSIRO top up scholarship.

# COMPETING INTEREST

None declared.